\documentclass{article}

\usepackage[preprint]{neurips_2019}

\usepackage[utf8]{inputenc} 
\usepackage[T1]{fontenc}    
\usepackage{hyperref}       
\usepackage{url}            
\usepackage{booktabs}       
\usepackage{amsfonts}       
\usepackage{nicefrac}       
\usepackage{microtype}      

\usepackage{xcolor}
\usepackage{graphicx}
\usepackage{amsmath}

\newcommand\fig[1]{Fig.~\ref{#1}}

\newcommand\tab[1]{Table~\ref{#1}}

\renewcommand{\vec}[1]{\boldsymbol{#1}}
\newcommand\mat[1]{\boldsymbol{\MakeUppercase{#1}}}

\title{On the difficulty of learning and predicting the long-term dynamics of bouncing objects}

\author{%
  Alberto Cenzato\thanks{Equal contribution, listing order is alphabetical.}\\
  Department of General Psychology\\
  University of Padova\\
  Padova, IT 35131 \\
  \texttt{alberto.cenzato@outlook.it} \\
   \And
  Alberto Testolin\footnotemark[1]\\
  Department of General Psychology and \\Department of Information Engineering\\
  University of Padova\\
  Padova, IT 35131 \\
  \texttt{alberto.testolin@unipd.it} \\
   \And
  Marco Zorzi\\
  Department of General Psychology and\\
  Padova Neuroscience Center\\
  University of Padova\\
  Padova, IT 35131 \\
  \texttt{marco.zorzi@unipd.it} \\
}

\begin{document}
\maketitle

\begin{abstract}
  The ability to accurately predict the surrounding environment is a foundational principle of intelligence in biological and artificial agents. In recent years, a variety of approaches have been proposed for learning to predict the physical dynamics of objects interacting in a visual scene. Here we conduct a systematic empirical evaluation of several state-of-the-art unsupervised deep learning models that are considered capable of learning the spatio-temporal structure of a popular dataset composed by synthetic videos of bouncing objects. We show that most of the models indeed obtain high accuracy on the standard benchmark of predicting the next frame of a sequence, and one of them even achieves state-of-the-art performance. However, all models fall short when probed with the more challenging task of generating multiple successive frames. Our results show that the ability to perform short-term predictions does not imply that the model has captured the underlying structure and dynamics of the visual environment, thereby calling for a careful rethinking of the metrics commonly adopted for evaluating temporal models. We also investigate whether the learning outcome could be affected by the use of curriculum-based teaching.
\end{abstract}

\section{Introduction} \label{sec:intro}

Learning the structure of event sequences has been a long-standing problem in neural network research \citep{elman1990finding,bengio1994learning,hochreiter1997long}, which has recently seen a resurgence thanks to the ground-breaking performance achieved by convolutional and recurrent deep networks \citep{graves2013speech,sutskever2014sequence,testolin2016learning}. Nevertheless, when it comes to model complex, high-dimensional sequences such as video streams, progress has been made mostly in the context of supervised learning, where the task requires to produce a series of labels or sentence-level natural language descriptions of the video content (e.g., \citep{donahue2015long}). There are, however, a variety of real-world scenarios where learning is completely unsupervised, and the goal is to build high-level internal representations of the environment that can be used to actively anticipate the content of the sensory stream \citep{testolin2016probabilistic}.

Here we specifically focus on the challenging task of video generation, where the goal is to learn the underlying dynamics of a complex visual stream in order to make accurate predictions of its temporal unfolding in the pixel-space. Several approaches have been proposed to learn how to predict frames from sequences of image patches \citep{srivastava2015unsupervised} or even real videos \citep{denton2017unsupervised,finn2016unsupervised,mathieu2015deep,vondrick2016generating}. However, capturing the complexity of the spatio-temporal structure embedded in real-world scenes turns out to be still out of reach even for state-of-the-art models, which are indeed usually tested only on short-term prediction tasks.

In this paper we thus consider a simpler setting, which involves the generation of synthetic videos whose content and dynamics can be more easily controlled and assessed. In particular, we employ the well-known \textit{bouncing balls} dataset originally introduced by \citet{sutskever2009recurrent}. This is perfectly suited as a benchmark because it removes most of the finer-grained visual complexity inherent in real-world scenes while retaining the rigid body physical dynamics that should be captured by learning. Predicting the long-term behavior of these moving objects is challenging, because even small prediction errors in the pixel space exponentially amplify as predictions are made deeper into the future.

We compare the long-term prediction performance of several state-of-the-art models, which have achieved good short-term prediction accuracy on the bouncing balls dataset or even on more complex, realistic video datasets. Surprisingly, it turns out that none of the tested models can reach satisfactory performance in long-term prediction, suggesting that unsupervised learning of good representations is harder than expected even for these simple synthetic videos. We also carry out an empirical assessment of curriculum-based learning strategies, aiming for a possible improvement in the long-term generation capability of the models.

\section{Background} \label{sec:background}

The bouncing balls dataset considered here consists of synthetic videos of $3$ white balls bouncing within a constrained box of black background, without friction nor gravity. In the original dataset each frame was of size $30 \times 30$ pixels, but we up-sampled the videos to $60 \times 60$ pixels to have a more reasonable image size. A series of frames from a sample sequence is shown in the top row of \fig{fig:one_step_sequence}.

The original approach for modeling the bouncing balls dataset was based on a probabilistic generative model, extending the restricted Boltzmann machine to the sequential domain \citep{sutskever2009recurrent}. This was one of the earliest, successful attempts in capturing high-order structure from pixel-level videos using an unsupervised learning model encoding contextual information in a distributed form. However, the spontaneous generation of video sequences turned out to be imprecise, resulting in a physical dynamics somehow more compatible with a Brownian than a Newtonian motion\footnote{Generated samples can be found in the Supplement of the original work: \href{https://papers.nips.cc/paper/3567-the-recurrent-temporal-restricted-boltzmann-machine-supplemental.zip}{https://papers.nips.cc/}.}. A step forward was proposed in \citet{gan2015deep}, where a deep, dynamic generative model was built using a sigmoid belief network that used a directed graph to generate sequences efficiently. The authors reported improved accuracy on the one-step-ahead prediction task.

A different approach was adopted by \citet{srivastava2015unsupervised}, which exploited a Long-Short Term Memory (LSTM) encoder-decoder network to learn both synthetic and realistic video sequences. Though the authors did not test their model on the bouncing balls videos, they evaluated its predictive ability using a somewhat similar synthetic dataset, where two handwritten digits moved (without collision) within a $64 \times 64$ pixels black background. The advantage of LSTMs over more traditional recurrent architectures lies on their ability to selectively retain temporal information depending on the input structure, which should be a critical feature to generate accurate long-term predictions also for the bouncing balls sequences.

Finally, inspired by modern theories of cortical computation \citep{friston2010free,clark2013whatever}, recent studies have tried to capture the complexity of video streams using predictive coding mechanisms \citep{lotter2015unsupervised,lotter2016deep,oord2018representation}. In particular, \citet{lotter2015unsupervised} reported state-of-the-art performance on the bouncing balls dataset in the one-step-ahead prediction task.

For the sake of completeness, we should also mention the existence of other modeling approaches placing stronger emphasis on explicit representations in learning and processing of structured information \citep{probmods2}. Though such framework has been used for learning physical dynamics from synthetic videos \citep{chang2017compositional}, it assumes that the model is inherently endowed with some generic knowledge of objects and their interactions, and the main purpose of learning is “just” to efficiently adapt to the specific properties observed in different environments. Our focus here is instead on models where no \textit{a priori} knowledge is built into the learning agent.

\section{Models} \label{sec:models}

In this section we provide a brief description of all learning architectures tested in our experiments. Most of them are a straightforward implementation of previously published work, while others include some novel adaptations. We provide a formal characterization to highlight the main features of each model; for an exhaustive description of previously published models the reader is referred to the original articles. All models have been re-implemented in PyTorch to ensure a consistent comparison and the complete source code is freely available\footnote{Our source code is freely available on \textcolor{blue}{\href{https://github.com/AlbertoCenzato/dnn_bouncing_balls}{GitHub}}, and it includes a script to generate the bouncing balls dataset (adapted from \citet{sutskever2009recurrent}).}.

\paragraph{LSTM} All models considered here are based on LSTM networks (\citet{hochreiter1997long}) so we use LSTMs both as a baseline model and for establishing a common notation that will be used throughout the paper. The LSTM used here has \textit{input} ($\vec{i}_t$), \textit{forget} ($\vec{f}_t$), and \textit{output} ($\vec{o}_t$) gates with no peephole connections; its architecture is summarized in the following equations:

    \begin{align*}
        \vec{i}_t & = \sigma(\mat{w}_{ii}\vec{x}_t + \vec{b}_{ii} + \mat{w}_{hi}\vec{h}_{(t-1)} + \vec{b}_{hi}) \\
        \vec{f}_t & = \sigma(\mat{w}_{if}\vec{x}_t + \vec{b}_{if} + \mat{w}_{hf}\vec{h}_{(t-1)} + \vec{b}_{hf}) \\
        \vec{o}_t & = \sigma(\mat{w}_{io}\vec{x}_t + \vec{b}_{io} + \mat{w}_{ho}\vec{h}_{(t-1)} + \vec{b}_{ho}) \\
        \vec{g}_t & = \tanh(\mat{w}_{ig}\vec{x}_t + \vec{b}_{ig} + \mat{w}_{hg}\vec{h}_{(t-1)} + \vec{b}_{hg})  \\
        \vec{c}_t & = \vec{x}_t \odot \vec{c}_{(t-1)} + \vec{i}_t \odot \vec{g}_t  \\
        \vec{h}_t & = \vec{o}_t \odot \tanh(\vec{c}_t)  \\
    \end{align*}

where $\sigma$ is the logistic sigmoid activation function, $\mat{w}\vec{x}$ and $\mat{w}\vec{h}$ are matrix-vector multiplications and $\odot$ is vector element-wise multiplication. The LSTM retains a memory of its past inputs in the state vectors $\vec{h}_t$ and $\vec{c}_t$, which are respectively the hidden and cell states of the network. The first one is the LSTM cell output which, in the case of stacked networks, is propagated as input to the next LSTM. $\vec{c}_t$ instead is the internal memory storage of the cell; the gates regulate how much of the current input to add to this state and what to remove from it. Finally $\vec{g}_t$ is a non-linear transformation of the input, whose addition to the cell state is regulated by the input gate $\vec{i}_t$.
The $60 \times 60 \times 1$ grayscale input frames $\mat{x}_0$, $\mat{x}_1$,~\dots, $\mat{x}_t$ are flattened into a sequence of vectors of size $3600$: $\vec{x}_0$, $\vec{x}_1$,~\dots, $\vec{x}_t$. The input sequence is fed into a stack of LSTM layers, which outputs the predicted next frame $\vec{\hat{x}}_{t+1}$. To predict $k$ frames ahead, each output $\vec{\hat{x}}_{t+i}$ is fed back as input to the LSTM to obtain $\vec{\hat{x}}_{t+i+1}$ until $\vec{\hat{x}}_{t+k}$ is reached.

\paragraph{ConvLSTM} First proposed by \citet{xingjian2015convolutional}, ConvLSTMs are designed to improve vanilla LSTM's ability in dealing with spatio-temporal data. In ConvLSTMs the matrix multiplications of the LSTM cell, which account for a fully connected gate activation, are replaced with a convolution over multiple channels. This means that hidden ($h_t$) and cell ($c_t$) states become tensors of shape $60 \times 60 \times \verb|number of channels|$. Prediction works as described for LSTM, but instead of using $\vec{x}_i$ as input vectors we use the $\mat{x}_i$ frames. The use of convolutional rather than fully connected gates gives a spatial prior that encourages the network to learn moving features detectors using less parameters. The LSTM equations reported above remain the same, except for the replacement of matrix-vector multiplications with convolutional operations (e.g., $\mat{w}_{ii}\vec{x}_t$ becomes $\mat{w}_{ii} \star \vec{x}_t$).

\begin{figure}
  \centering
  \includegraphics[width=0.7\linewidth]{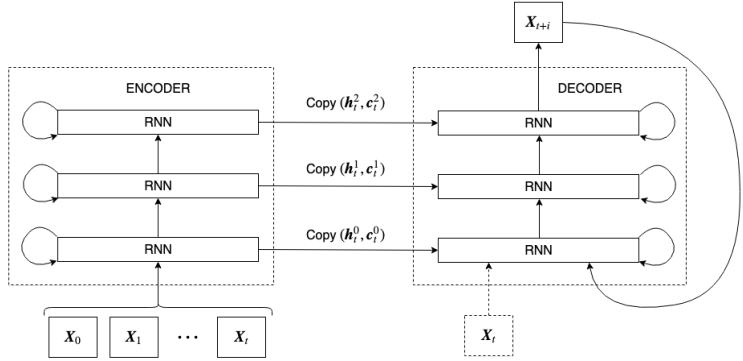}
  \caption{Architecture of an autoregressive sequence-to-sequence model with generic RNN layers; in this work they are either LSTM or ConvLSTM layers, so we copy the hidden states $(\vec{h}_t^l, \vec{c}_t^l)$, but in principle one could use any RNN.}
  \label{fig:seq2seq}
\end{figure}

\paragraph{Seq2seq ConvLSTM} Predicting future video frames can be cast as a sequence-to-sequence mapping problem: given the frames $\mat{x}_0$, $\mat{x}_1$,~\dots, $\mat{x}_t$ as input sequence, we want to predict $\mat{x}_{t+1}$, $\mat{x}_{t+2}$,~\dots, $\mat{x}_{t+k}$ as output sequence. The two sequences do not necessarily have the same length: in a long-term generation setting, we might want the output sequence to be as long as possible, while providing a short input sequence (i.e., the context). \citet{sutskever2014sequence} proposed an effective sequence-to-sequence model that maps an input sequence into a fixed-size vector, and then uses this vector to produce the output sequence. This model was first introduced in the context of natural language processing, but can be adapted to image sequences. As in \citet{srivastava2015unsupervised}, we implemented it as a stack of LSTMs which map the flattened video frames $\vec{x}_0$, $\vec{x}_1$,~\dots, $\vec{x}_t$ into the vectors $(\vec{h}_t^l, \vec{c}_t^l)$, which are respectively the hidden and cell states of the $l^{th}$ LSTM layer. The next frame $\vec{x}_{t+1}$ can be predicted starting from the state $(\vec{h}_t^l, \vec{c}_t^l)$ and giving $\vec{x}_t$ as input to another LSTM stack; for multi-steps predictions, $\vec{\hat{x}}_{t+i}$ is then fed back until $\vec{\hat{x}}_{t+k}$ is reached\footnote{Having two different LSTM stacks serves two purposes. First, the network is forced to learn two functions: an \textit{encoder} function $f$ that projects the input sequence into vectors by extracting the relevant spatio-temporal features, and a \textit{decoder} function $g$ that uses these vectors to produce output sequences; in general, these two could be arbitrarily different functions. A monolithic model would be required to learn the composite function $h = f \circ g$, which might be hard. Second, encoding a sequence into a single vector forces the model to find an effective, compressed state representation that retains as much information as possible.}. A diagram of the model architecture is shown in \fig{fig:seq2seq}.

Due to hardware limitations, we could not test the seq2seq architecture using fully-connected LSTM layers, so we opted for an architecture with fewer parameters. As proposed by \citet{xingjian2015convolutional}, the seq2seq ConvLSTM uses an architecture very similar to the seq2seq LSTM of \citet{sutskever2014sequence}, with ConvLSTM layers instead of LSTMs and an additional $1 \times 1$ convolutional layer that receives the concatenation of all decoder's channels and outputs the predicted image. Our implementation removes this last layer, and uses the architecture shown in \fig{fig:seq2seq}. Having ConvLSTM layers there is no need of flattening the input frames, therefore we used the unflattened $60 \times 60 \times 1$ video frames $\mat{x}_0$, $\mat{x}_1$,~\dots, $\mat{x}_t$ as input sequence, and the hidden state vectors $\vec{h}_t^l$ and $\vec{c}_t^l$ become tensors of size $60 \times 60 \times \verb|number of channels|$.

\paragraph{Multi-decoder seq2seq ConvLSTM} Building on top of the seq2seq LSTM model, \citet{srivastava2015unsupervised} attached multiple decoders on the same encoder allowing multiple output sequences for the same input sequence. Carefully choosing decoders' tasks to be different, but not orthogonal, forces the encoder to learn a common representation with features that can be used by all decoders. As \citet{srivastava2015unsupervised} we used two decoders, one to reconstruct the reversed input sequence and the other to predict future frames; both tasks can be accomplished exploiting the same information: position, speed and direction of the balls at the end of the input sequence. Training the encoder together with two decoders should encourage the encoder to learn a representation that does not simply memorize the sequence, but still contains such information. We tried to improve the model of \citet{srivastava2015unsupervised} by using ConvLSTM layers instead of LSTM layers; this should maintain the advantages of seq2seq and multi-decoder models, while adding the spatio-temporal capabilities of ConvLSTM. Multi-decoders aside, the implementation is the same of seq2seq ConvLSTM.

\paragraph{RTRBM, DTSBN and predictive coding models} As benchmarks, we also compare our results with the one-step prediction accuracy previously reported for the Recurrent Temporal Restricted Boltzmann Machine \citep{sutskever2009recurrent}, the Deep Temporal Sigmoid Belief Network \citep{gan2015deep} and the predictive coding model of \citet{lotter2015unsupervised}.

\section{Training and testing methods} \label{sec:train}
The best hyperparameters for each model were established using a grid-like search procedure, varying one parameter at a time and keeping the others fixed; for each architecture we varied the number of layers, hidden units and/or channels, learning rate, number of epochs and batch size\footnote{Each hyperparameter was varied independently, because performing a full grid-search was unfeasible due to the computational complexity of the models considered. See Supplementary Material for a detailed explanation of the parameters ranges and optimization results.}. All models were trained on a training dataset with $6000$ sequences of $40$ frames each; hyperparameters optimization was performed on a validation dataset composed by $1200$ videos, while final evaluation was carried out on a separate test set of $1200$ videos.

The best configuration resulting for each model was then trained using two strategies. In the simplest one, the model receives in input the frame $\mat{x}_i$ and predicts $\mat{\hat{x}}_{i+1}$ at each time step $i$; this procedure is called \textit{teacher forcing} because the new input is always \textit{given}, and does not depend upon previous model predictions. One main issue with teacher forcing is that the model only observes sample frames from the training dataset; during long-term generation, though, it must autoregressively predict $k$ frames ahead and therefore will receive its own predictions $\mat{\hat{x}}_{t+i}$ as input from frame $t+1$ to frame $t+k$. Such input is not drawn from the training dataset, but from the model's frame-generating distribution; the model does not expect possible errors in $\mat{\hat{x}}_{t+i}$, which quickly sum up producing increasingly worse predictions at each frame.

One way to deal with the mismatch between the training and generated distributions is to train the model on its own predictions. During training, after $t$ ground-truth context frames, the model is given its own subsequent predictions $\mat{\hat{x}}_{t+i}$ as input; error is then computed, as for teacher forcing, in terms of the difference between the predicted and ground-truth sequences: 
$$\frac{1}{k}\sum_{i=1}^k L(\mat{\hat{x}}_{t+i}, \mat{x}_{t+i})$$
where $L(\cdot)$ is the loss function. However, since at early learning stages it might be hard for the model to learn in this ``blind'' prediction mode\footnote{We call it ``blind'' because the model, when predicting frames $\mat{\hat{x}}_{t+i}$ does not have any information about the ground-truth frames $\mat{x}_{t+i}$.} we also explored a \textit{curriculum training} routine inspired by \citet{zaremba2014learning}. Our curriculum strategy progressively blends teacher forcing into blind training, by increasing the task difficulty as the model becomes more and more accurate during learning. Training starts in a simple teacher forcing setting, where $\mat{x}_{t+i}$ is used as input to predict frames $\mat{x}_{t+i+1}$, ; after $10$ epochs the task difficulty is raised and to predict one of the $\mat{x}_{t+i+1}$ frames the previously predicted frame $\mat{\hat{x}}_{t+i}$ is used. So $1$ out of $k$ frames is generated using the model's own prediction as input. Task difficulty is increased every $10$ epochs, each time using one additional predicted frame as input. We hypothesized that curriculum training would facilitate long-term predictions, being the model gradually accustomed to perform predictions on its own generated frames.

The evaluation of model performance is usually carried out by measuring the Mean Squared Error (MSE) between predicted and ground-truth frames, since this is the most commonly used loss function during training. In the bouncing balls dataset motion is deterministic, therefore we introduce an additional evaluation metric based on the distance between the centroid of the $j^{th}$ predicted ball ($\vec{\hat{c}}_j$) and the centroid of its nearest neighbor ball in the ground-truth frame ($\vec{c}_i$):
$$d(\mat{\hat{f}}_{t+i}, \mat{f}_{t+i}) = \sum_{j=0}^{\hat{n}-1} ||\vec{\hat{c}}_j - \vec{c}_j||_2 + |\hat{n}_{t+i} - n_{t+i}| \cdot 60\sqrt{2}$$
where $\hat{n}_{t+i}$ is the number of balls in $\mat{\hat{f}}_{t+i}$ and $n_{t+i}$ is the number of balls in $\mat{f}_{t+i}$. The last term of the equation adds a penalty to $d(\cdot)$ when there is a mismatch between $\hat{n}_{t+i}$ and $n_{t+i}$; $60\sqrt{2}$ was chosen because it is the diagonal length of the $60 \times 60$ squared frame which is more than the maximum distance between two balls. Simple computer vision algorithms were used to determine the coordinate of balls in each frame\footnote{We compute the centroids of the connected components in each frame of the sequence, by first eroding each image and removing pixels under a fixed threshold. Erosion and threshold are applied to avoid that two colliding balls are detected as one single object.}. Both MSE and centroid distance are measured for each frame, and then averaged over all sequences of the testing dataset. To maintain compatibility between our results and the MSE values reported by previous research performed on smaller frame sizes (i.e., $30 \times 30$ pixels), our MSE values were scaled by a factor of $4$.

All the models were trained on a desktop PC equipped with a Nvidia GTX 1080 GPU with 8GB of dedicated memory. Training time, depending on the model, varied between 1 and 12 hours. Image pre-processing was kept at a minimum: input frames were only normalized in $[-1, 1]$.

\section{Results} \label{sec:results}
In this section we present the results obtained with all models considered in Sec. \ref{sec:models}, trained using the best configuration resulting from the hyperparameters optimization procedure.

\begin{figure}
  \includegraphics[width=0.78\linewidth]{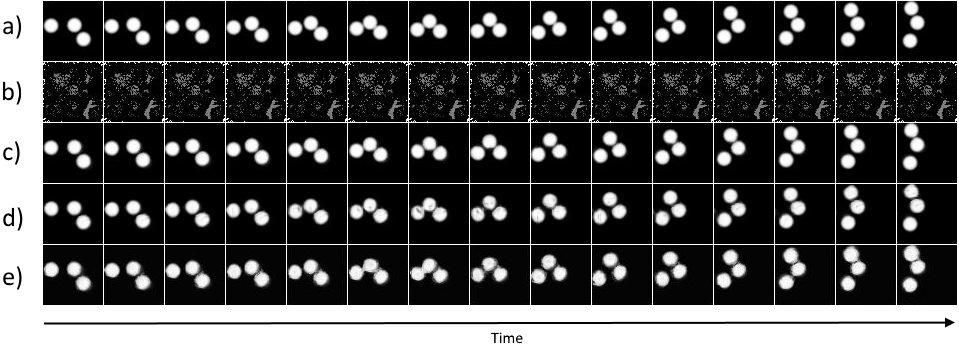}
  \caption{Samples of one-step-ahead predictions from all the models considered: (a) ground-truth sequence, (b) LSTM, (c) ConvLSTM, (d) seq2seq ConvLSTM, (e) seq2seq ConvLSTM multi-decoder.}
  \label{fig:one_step_sequence}
\end{figure}

As shown in \fig{fig:one_step_sequence}, in the one-step prediction task all models (apart for LSTM) are able to accurately predict the next frame, given $10$ frames as context. \tab{tab:one_step_errors} reports both average MSE and centroid distances: even the simple ConvLSTM achieves excellent performance, and more complex models do not exhibit significantly improved prediction. As expected, the centroid distance metric is in good alignment with MSE. Vanilla LSTM is the only model unable to accurately perform one-step predictions, probably due to the high number of parameters and lack of explicit spatial biases: indeed, after $30$ epochs learning did not yet converge and was stopped. Crucially, in the one-step-ahead prediction task all convolutional models are well aligned with the performance reported in previous studies: in fact, the ConvLSTM accuracy is even higher than the current state-of-the-art reported by \citet{lotter2015unsupervised}.

\begin{figure}
  \includegraphics[width=0.98\linewidth]{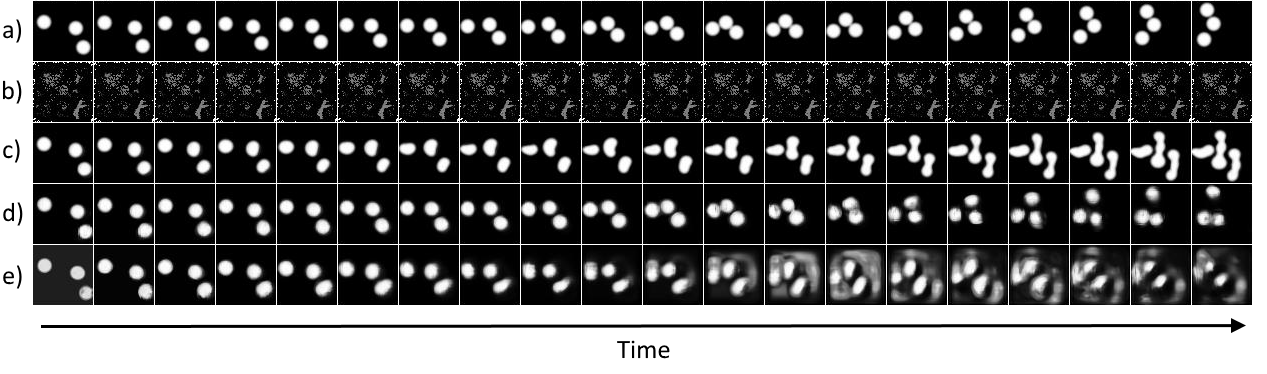}
  \caption{Samples of 20-steps-ahead predictions from all the models considered: (a) ground-truth sequence, (b) LSTM, (c) ConvLSTM, (d) seq2seq ConvLSTM, (e) seq2seq ConvLSTM multi-decoder.}
  \label{fig:20_step_sequence}
\end{figure}

However, accurate one-step-ahead prediction does not entail that the model has captured the underlying spatio-temporal dynamics of the videos. This phenomenon is clearly visible in \fig{fig:20_step_sequence}, which shows samples of generated sequences in a more challenging setting, where the next $20$ frames have to be predicted given a context of $10$ frames. The striking result is that most of the models cannot maintain the visual dynamics for more than just a few frames. A common misprediction pattern consists in expanding the balls into larger blobs, as evident from the ConvLSTM generation: the balls are increasingly stretched and loose their shape. From a qualitative point of view, it seems that ConvLSTM distorts the objects but is able to maintain an overall sharp appearance for the distribution of active pixels. This might explain why it achieves the lowest error in the one-step-ahead prediction task: even if the trajectory dynamics is not captured at all, the spatial consistency between early consecutive frames is almost perfect. The seq2seq ConvLSTM is clearly the model that generates video sequences with the best visual appearance, though object dynamics seems to diverge from ground-truth after about $15$ predicted frames. Additional examples of generated sequences for each model (with a 200-step-ahead generation range) are included in the source code folder.

\begin{figure}
    \centering
    \includegraphics[width=0.85\linewidth]{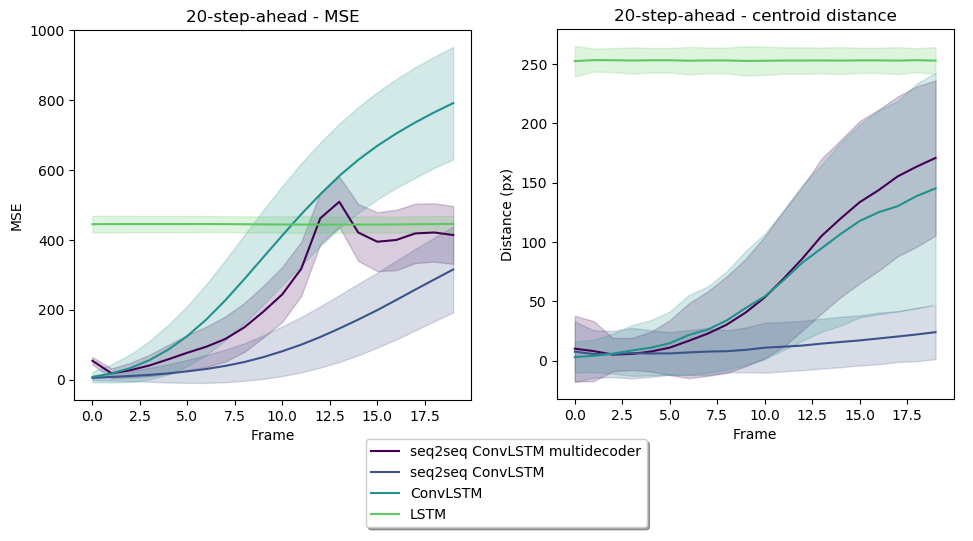}
    \caption{Plots showing how prediction errors increase as frame generations are made deeper into the future. The shaded area around the lines corresponds to the standard deviation of the mean error.}
    \label{fig:20_step_cumulative_error}
\end{figure}

Average MSE and centroid distances are reported in \tab{tab:20_step_errors}: ConvLSTM are indeed much less accurate in the long-range prediction task, and the best performance is achieved by sequence-to-sequence architectures. \fig{fig:20_step_cumulative_error} shows how errors rapidly accumulate when predicting the $20$ frames (the LSTM error has been included as a semi-random baseline). It is worth noting that the seq2seq ConvLSTM obtains the best results both in terms of MSE and centroid distance; its prediction accuracy, in particular with respect to centroid distance, degrades gently as time passes, maintaining the best trajectory estimation. Interestingly, a qualitative analysis on the internal encoding of seq2seq architectures revealed how these models might encode the input sequence into a static state vector by creating a spatial representation of object trajectories (see Supplementary Material for further discussion).

\begin{table}
  \caption{One-step-ahead prediction error (mean and standard error) for all models in terms of MSE and centroid distance in pixels. The latter is omitted for LSTM because the predicted frames are mostly filled with noise. As baseline we report the error of a null model that always predicts empty frames.}
  \label{tab:one_step_errors}
  \centering
  \begin{tabular}{lrrr}
    \toprule
    Model                          & MSE           & Centroid distance (px) \\
    \midrule
    LSTM                           & $111.09 \pm 0.68$ & n.a.            \\
    ConvLSTM                       & $0.58   \pm 0.16$ & $1.84 \pm 0.32$ \\ 
    seq2seq ConvLSTM               & $1.34   \pm 0.19$ & $7.88 \pm 0.76$ \\
    seq2seq ConvLSTM multi-decoder & $4.55   \pm 0.40$ & $6.80 \pm 0.64$ \\
    \midrule
    \citet{sutskever2009recurrent} & $3.88 \pm 0.33$   & -      \\
    \citet{gan2015deep}            & $2.79 \pm 0.39$   & -      \\
    \citet{lotter2015unsupervised} & $0.65 \pm 0.11$   & -      \\
    \midrule
    Baseline (empty frame)         & $93.07$           & $254,56$        \\
    \bottomrule
  \end{tabular}
\end{table}

\begin{table}
  \caption{Prediction errors for all models in the more challenging 20-step-ahead prediction task.}
  \label{tab:20_step_errors}
  \centering
  \begin{tabular}{lcrr}
    \toprule
    Model                          & MSE           & Centroid distance (px) \\
    \midrule
    LSTM                           & $111.11 \pm 0.67$ & n.a.              \\
    ConvLSTM                       & $95.67  \pm 3.31$ & $61.76 \pm 1.51$ \\  
    seq2seq ConvLSTM               & $27.13  \pm 1.91$ & $11.73 \pm 0.60$ \\
    seq2seq ConvLSTM multi-decoder & $60.37  \pm 1.79$ & $67.88 \pm 1.31$ \\
    \bottomrule
  \end{tabular}
\end{table}

We finally investigated whether the curriculum teaching regimen, where ground-truth frames are provided during learning as prediction unfolds over time, would improve the generation performance. However, as reported in the Supplementary Material, we did not observe a significant improvement in terms of metrics and/or visual appearance of the generated sequences.

\section{Conclusions} \label{sec:conclusion}
In this paper we presented an extensive empirical evaluation of several unsupervised deep learning architectures, considered among the state-of-the-art for modeling complex time series involving high-dimensional data. We focused our investigation on a popular dataset containing thousands of synthetic videos of bouncing balls, which has traditionally been used as a test bed in this domain.

In agreement with previous studies, our simulations demonstrate that different approaches indeed achieve good performance on a standard one-step-ahead prediction task. However, long-term prediction proved much harder than expected: all models considered, including those reporting state-of-the-art in the standard prediction task, fall short when probed on the task of generating multiple successive frames, suggesting that none of the models successfully captured the underlying physical dynamics of the videos.

It should be stressed that our results do not demonstrate that these models \textit{cannot} learn the required task: computational complexity did not allow us to implement an exhaustive hyperparameter optimization, so one could still argue that some of the models might in principle succeed, if properly parameterized. However, our simulations clearly demonstrate the difficulty in finding proper configurations, suggesting that the robustness of these models should be better explored.

Another crucial factor that should be considered is the loss function used to train the models. Previous research \citep{ranzato2014video, mathieu2015deep, srivastava2015unsupervised} highlighted the limitations of pixel-level loss functions such as MSE or cross-entropy, which often produce blurry images since they are focused on per-pixel appearance: image structure is not explicitly taken into account, so a black background filled with salt-and-pepper noise might have the same loss of a nice-looking frame containing objects with slightly distorted shapes or small coordinates offset. To ensure compatibility with previous results in our work we used MSE loss, but alternatives based on adversarial loss or mutual information are under investigation (e.g., \citep{hjelm2018learning, oord2018representation})

Overall, our findings highlight the necessity to establish proper (and common) methodological practices to more carefully evaluate predictive models: achieving accurate performance in one-step prediction tasks does not necessarily imply that the model has captured the spatio-temporal structure in the data. Some recent approaches have started focusing on long-term prediction ability to overcome this critical limitation \citep{villegas2017learning}; one interesting future research direction would then be to compare the performance of machines with that of human observers, who are capable of tracking moving objects for a long time interval \citep{tresilian1995perceptual}, but also suffer from perceptual limitations when multiple objects are involved \citep{cavanagh2005tracking}.

\bibliography{bibliography}

\clearpage
\appendix

\section{Hyperparameters Optimization} \label{sec:optimization}

We carried out an extensive search on the hyperparameters space in order to find the configuration that could achieve higher performance for each model. Due to the high computational complexity of all models considered, hyperparameter optimization was carried out by varying one hyperparameter at a time and keeping the others fixed. For each architecture we varied the number of layers, hidden units or channels, learning rate, number of epochs and batch size. Hyperparameters optimizaton was performed on a validation dataset composed by $1200$ videos, while final evaluation was carried out on a separate testing dataset of the same size. In \tab{tab:hyper_range} we report the range of hyperparameters tested for each model.

\begin{figure}
    \centering
    \includegraphics[width=0.8\linewidth]{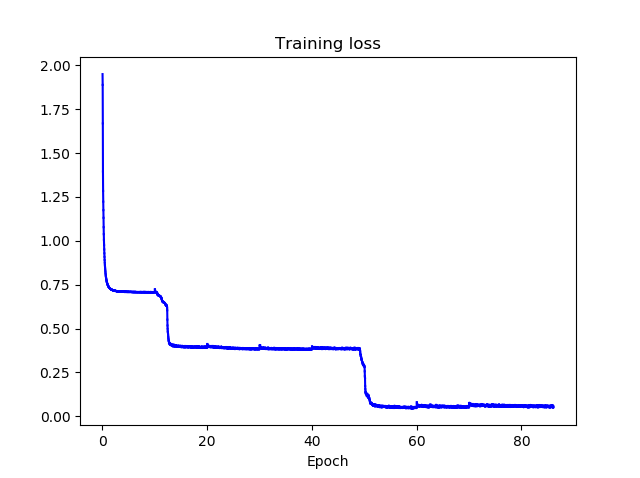}
    \caption{Example of a sequence-to-sequence ConvLSTM multi-decoder model trained with a curriculum regimen. The training loss highlights three plateaus/local minima. The small peaks visible after every $10$ epochs mark when task complexity increased due to the curriculum procedure.}
    \label{fig:local_minima}
\end{figure}

The bouncing balls dataset proved to be a particularly hard optimization problem. For many hyperparameter configurations the optimizer got stuck in plateaus or local minima; we show an example of the resulting training loss (MSE) in \fig{fig:local_minima}. Most of the time these plateaus / local minima were associated with minimum activation on all output units (i.e. empty, black frames). This issue might be specific for the bouncing balls dataset (or other videos with uniform background) and we would not expect to see it with more realistic video datasets. The final configurations obtained after hyperparameter search are shown in \tab{tab:best_params}.

\begin{table}
  \caption{Ranges of tested hyperparameters for each architecture. For sequence-to-sequence models we provide layers, kernel sizes and channels for the encoder only; decoders use the same values (i.e. encoders and decoders have the same architecture). ${}^*$ For LSTM each additional hidden layer had $1024$ hidden units with the exception of input layer, while for epochs we used the custom values $[30, 60, 100, 150, 200]$.}
  \label{tab:hyper_range}
  \centering
  \begin{tabular}{lccc}
    \toprule
    \textbf{param} & \textbf{min} & \textbf{max} & \textbf{step} \\
    \midrule
    \multicolumn{2}{c}{LSTM} \\
    \cmidrule(r){1-2}
    layers          & 1        & 5                            & 1            \\
    hidden units    & $[3600]$ & $[2048,1024,1024,1024,3600]$ & custom${}^*$ \\
    \\
    \multicolumn{2}{c}{ConvLSTM} \\
    \cmidrule(r){1-2}
    layers          & 1       & 5       & 1           \\
    learning rate   & 0.0001  & 0.1     & $\times 10$ \\
    kernel size     & $(3,3)$ & $(7,7)$ & $(2,2)$     \\
    channels        & 1       & 30      & 5           \\
    epochs          & 30      & 200     & $\sim 50^*$ \\
    minibatch       & 8       & 32      & $\times 2$  \\
    \\
    \multicolumn{2}{c}{Seq2seq ConvLSTM} \\
    \cmidrule(r){1-2}
    layers          & 1       & 5       & 1           \\
    learning rate   & 0.0001  & 0.1     & $\times 10$ \\
    kernel size     & (3,3)   & (7,7)   & (2,2)       \\
    channels        & 1       & 30      & 5           \\
    epochs          & 30      & 200     & $\sim 50^*$ \\
    minibatch       & 8       & 16      & $\times 2 $ \\
    \\
    \multicolumn{2}{c}{Seq2seq ConvLSTM multidecoder} \\
    \cmidrule(r){1-2}
    layers          & 1       & 5       & 1           \\
    learning rate   & 0.0001  & 0.1     & $\times 10$ \\
    kernel size     & (3,3)   & (7,7)   & (2,2)       \\
    channels        & 1       & 30      & 5           \\
    epochs          & 30      & 200     & $\sim 50^*$ \\
    minibatch       & 8       & 16      & $\times 2 $ \\
    \bottomrule
  \end{tabular}
\end{table}

\begin{table}
  \caption{Hyperparameters for the best models}
  \label{tab:best_params}
  \centering
  \begin{tabular}{lllll}
    \toprule
    Param                 & LSTM       & ConvLSTM    & s2s ConvLSTM   & s2s ConvLSTM multi \\
    \midrule
    training              & curriculum & forced      & forced         & curriculum      \\
    kernels               & -          & $[7,5,3,3]$ & $[7, 5, 3, 3]$ & $[7,5,3,3,3]$   \\
    channels              & -          & $[10,10,10,1]$ & $[10,10,10,1]$ & $[10, 10, 10, 10, 1]$ \\
    hidden units          & $[2048,1024,1024,3600]$ & - & - & - \\
    minibatch             & 16         & 16          & 16             & 16              \\
    epochs                & 30         & 100         & 100            & 100             \\
    learing rate          & 0.001      & 0.001       & 0.001          & 0.001           \\
    \bottomrule
  \end{tabular}
\end{table}

\section{Teacher forced \textit{vs}. Curriculum learning}
\label{sec:teacher_vs_curriculum}
\begin{figure}
    \centering
    \includegraphics[width=\linewidth]{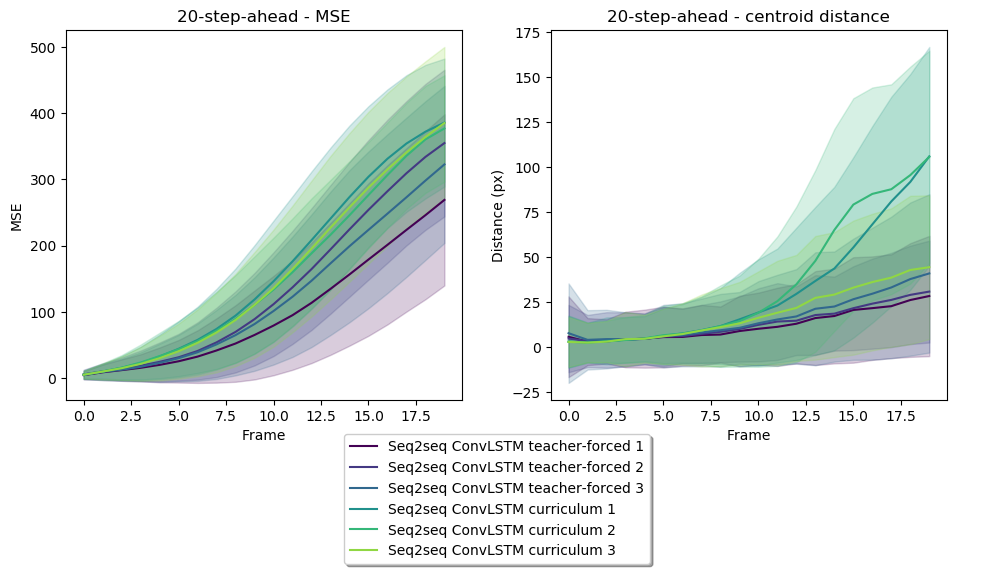}
    \caption{MSE and centroid distance evaluation for $6$ models trained with and without curriculum.}
    \label{fig:teacher_vs_curr_plot}
\end{figure}

\begin{figure}
    \centering
    \includegraphics[width=\linewidth]{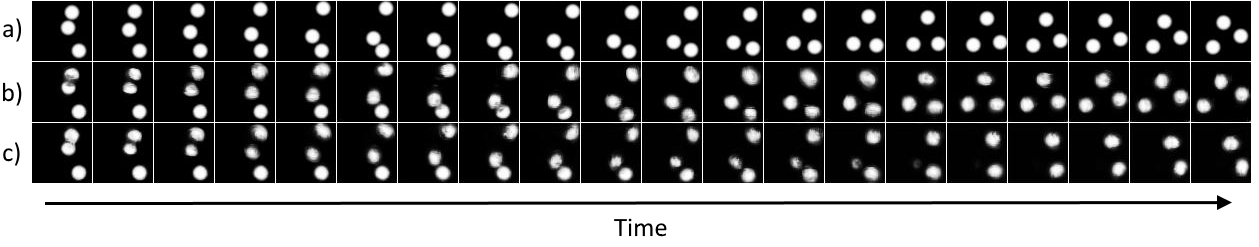}
    \caption{20-steps-ahead predictions generated by sequence-to-sequence ConvLSTM with best available hyperparameters. Comparison between (a) ground-truth, (b) teacher forced and (c) curriculum trained models.}
    \label{fig:teacher_vs_curr_sequence}
\end{figure}

One of the strategies pursued to improve prediction accuracy was to progressively increase task difficulty during training. The models were encouraged to make predictions further and further into the future as training epochs passed. Although we expected to see both MSE and centroid distance improve using curriculum learning, we obtained mixed results: some models achieved better performance under the curriculum regimen, but others did not (also see \tab{tab:best_params}). To further investigate the impact of curriculum-based strategies, we trained $6$ sequence-to-sequence ConvLSTM models with identical hyperparameters and architecture; half were trained with curriculum, half with teacher-forcing. The tests on one-step-ahead prediction did not provide clear results; on the contrary $20$-step prediction stressed the differences between the $6$ models. As shown in \fig{fig:teacher_vs_curr_plot}, it seems that curriculum learning affects both MSE and centroid distance measures: while curriculum-trained models have similar accuracy to non-curriculum-trained models in terms of MSE, trajectory estimation accuracy drops for most of these models. Thus curriculum learning seems to be \textit{at most} as good as teacher-forcing, but with higher model variability and at the cost of a more complex training procedure. An example of a $20$ frames prediction for the two best models with curriculum and teacher-forced training is shown in \fig{fig:teacher_vs_curr_sequence}; notice how in the curriculum sequence one ball fades away and the others move in the wrong direction.

\section{Spatial coding of temporal structure in sequence-to-sequence models}
\label{sec:seq2seq_encoding}

One of the most interesting features of sequence-to-sequence models is their ability to encode a temporal sequence into a fixed-size vector, which should retain as much as possible of the input spatio-temporal structure to allow accurate generations during the sequential decoding phase. We tried to investigate how temporal information was encoded in this type of architectures by visualizing the channel activations corresponding to a specific input sequence.

\begin{figure}[hb]
    \centering
    \includegraphics[width=\linewidth]{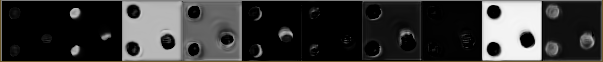}
    \caption{Hidden state representation of the $3^{rd}$ encoding layer for a sequence-to-sequence ConvLSTM multi-decoder model.}
    \label{fig:hidden_representation_bad}
\end{figure}

\begin{figure}[hb]
    \centering
    \includegraphics[width=0.25\linewidth]{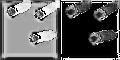}
    \caption{Hidden state representation of the last encoding layer for a multi-decoder model with reduced number of channels, smaller kernels ($3\times 3$) and only this hidden state as bottleneck between encoder and decoder.}
    \label{fig:hidden_representation_good}
\end{figure}

Interestingly, we can observe how the model internally uses a form of spatial encoding to capture the temporal structure, for example by producing a fading white edge in the motion direction of each ball. In \fig{fig:hidden_representation_bad} we show an example of internal encoding for the $10$ channels $\vec{h}_t^{3}$ of a sequence-to-sequence ConvLSTM multi-decoder model: channels $5$ and $7$ seems to exhibit a preference for the motion direction. A more striking form of encoding, though, can be obtained with a modified version of this model, where the encoder is forced to compress the whole input sequence in a tensor with only $2$ channels. As shown in \fig{fig:hidden_representation_good}, in this case the channels encode the complete trajectory of the balls in the input sequence.

\end{document}